\documentclass[sigconf]{acmart}
\AtBeginDocument{%
  \providecommand\BibTeX{{%
    \normalfont B\kern-0.5em{\scshape i\kern-0.25em b}\kern-0.8em\TeX}}}

\renewcommand\footnotetextcopyrightpermission[1]{}
\usepackage{bm}
\usepackage{wasysym}
\usepackage{makecell}
\usepackage{multirow}
\begin{document}

%%
%% The "title" command has an optional parameter,
%% allowing the author to define a "short title" to be used in page headers.
\title{DiffStereo: High-Frequency Aware Diffusion Model for Stereo Image Restoration}

%%
%% The "author" command and its associated commands are used to define
%% the authors and their affiliations.
%% Of note is the shared affiliation of the first two authors, and the
%% "authornote" and "authornotemark" commands
%% used to denote shared contribution to the research.
% \author{Ben Trovato}
% \authornote{Both authors contributed equally to this research.}
% \email{trovato@corporation.com}
% \orcid{1234-5678-9012}
% \author{G.K.M. Tobin}
% \authornotemark[1]
% \email{webmaster@marysville-ohio.com}
% \affiliation{%
%   \institution{Institute for Clarity in Documentation}
%   \streetaddress{P.O. Box 1212}
%   \city{Dublin}
%   \state{Ohio}
%   \country{USA}
%   \postcode{43017-6221}
% }
\author{Huiyun Cao}
\authornote{Both authors contributed equally to this research.}
\affiliation{%
  \institution{Shenzhen International Graduate School, Tsinghua University}
  \city{Shenzhen}
  \country{China}}
\email{caohy22@mails.tsinghua.edu.cn}

\author{Yuan Shi}
\authornotemark[1]
\affiliation{%
  \institution{Shenzhen International Graduate School, Tsinghua University}
  \city{Shenzhen}
  \country{China}}
\email{shiy22@mails.tsinghua.edu.cn}

\author{Bin Xia}
\affiliation{%
  \institution{Department of Computer Science and Engineering, Chinese University of Hong Kong}
  \city{Hongkong}
  \country{China}}
\email{zjbinxia@gmail.com}

\author{Xiaoyu Jin}
\affiliation{%
  \institution{Shenzhen International Graduate School, Tsinghua University}
  \city{Shenzhen}
  \country{China}}
\email{jinxy22@mails.tsinghua.edu.cn}

\author{Wenming Yang}
\affiliation{%
  \institution{Shenzhen International Graduate School, Tsinghua University}
  \city{Shenzhen}
  \country{China}}
\email{yang.wenming@sz.tsinghua.edu.cn}
\authornote{Corresponding author.}

% %% Author affiliation
% \affiliation[a]{organization={Shenzhen International Graduate School, Tsinghua University},%Department and Organization
%             city={Shenzhen}, 
%             postcode={518055}, 
%             country={China}}
% \affiliation[b]{organization={Department of Computer Science and Engineering, Chinese University of Hong Kong},%Department and Organization
%             city={Hong Kong}, 
%             postcode={999077},
%             country={China}}
%% You do not have to enter your paper ID

%%
%% By default, the full list of authors will be used in the page
%% headers. Often, this list is too long, and will overlap
%% other information printed in the page headers. This command allows
%% the author to define a more concise list
%% of authors' names for this purpose.
% \renewcommand{\shortauthors}{Trovato and Tobin, et al.}

%%
%% The abstract is a short summary of the work to be presented in the
%% article.
\begin{abstract}
Diffusion models (DMs) have achieved promising performance in image restoration but haven’t been explored for stereo images. The application of DM in stereo image restoration is confronted with a series of challenges. The need to reconstruct two images exacerbates DM’s computational cost. Additionally, existing latent DMs usually focus on semantic information and remove high-frequency details as redundancy during latent compression, which is precisely what matters for image restoration. To address the above problems, we propose a high-frequency aware diffusion model, DiffStereo for stereo image restoration as the first attempt at DM in this domain. Specifically, DiffStereo first learns latent high-frequency representations (LHFR) of HQ images. DM is then trained in the learned space to estimate LHFR for stereo images, which are fused into a transformer-based stereo image restoration network providing beneficial high-frequency information of corresponding HQ images. The resolution of LHFR is kept the same as input images, which preserves the inherent texture from distortion. And the compression in channels alleviates the computational burden of DM. Furthermore, we devise a position encoding scheme when integrating the LHFR into the restoration network, enabling distinctive guidance in different depths of the restoration network. Comprehensive experiments verify that by combining generative DM and transformer, DiffStereo achieves both higher reconstruction accuracy and better perceptual quality on stereo super-resolution, deblurring, and low-light enhancement compared with state-of-the-art methods.
\end{abstract}

% Diffusion models (DMs) have achieved promising performance in xxx but the application to xxx remains unexplored. One aspect of the issue is the significant computational costs incurred by DMs, which are further amplified when reconstructing two images. Additionally, traditional DM-based methods tend to generate undesired details, which are perceptually high-quality but with poor fidelity.

%%
%% The code below is generated by the tool at http://dl.acm.org/ccs.cfm.
%% Please copy and paste the code instead of the example below.
%%
\begin{CCSXML}
<ccs2012>
 <concept>
  <concept_id>00000000.0000000.0000000</concept_id>
  <concept_desc>Do Not Use This Code, Generate the Correct Terms for Your Paper</concept_desc>
  <concept_significance>500</concept_significance>
 </concept>
 <concept>
  <concept_id>00000000.00000000.00000000</concept_id>
  <concept_desc>Do Not Use This Code, Generate the Correct Terms for Your Paper</concept_desc>
  <concept_significance>300</concept_significance>
 </concept>
 <concept>
  <concept_id>00000000.00000000.00000000</concept_id>
  <concept_desc>Do Not Use This Code, Generate the Correct Terms for Your Paper</concept_desc>
  <concept_significance>100</concept_significance>
 </concept>
 <concept>
  <concept_id>00000000.00000000.00000000</concept_id>
  <concept_desc>Do Not Use This Code, Generate the Correct Terms for Your Paper</concept_desc>
  <concept_significance>100</concept_significance>
 </concept>
</ccs2012>
\end{CCSXML}

\ccsdesc[500]{Computing methodologies → Reconstruction; Matching.}

%%
%% Keywords. The author(s) should pick words that accurately describe
%% the work being presented. Separate the keywords with commas.

\keywords{Stereo Image Restoration, Diffusion Model, Transformer}
%% A "teaser" image appears between the author and affiliation
%% information and the body of the document, and typically spans the
%% page.
% \begin{teaserfigure}
%   \includegraphics[width=\textwidth]{sampleteaser}
%   \caption{Seattle Mariners at Spring Training, 2010.}
%   \Description{Enjoying the baseball game from the third-base
%   seats. Ichiro Suzuki preparing to bat.}
%   \label{fig:teaser}
% \end{teaserfigure}

% \received{20 February 2007}
% \received[revised]{12 March 2009}
% \received[accepted]{5 June 2009}

%%
%% This command processes the author and affiliation and title
%% information and builds the first part of the formatted document.
\maketitle

\section{Introduction}
Stereoscopic imaging systems shoot a scene from left and right perspectives. According to the geometric relationship in a stereo pair, the parallax between the two views and the depth of objects can be calculated and further we can analyze the 3D structure of the scene. Stereo cameras and stereo images have been extensively used in diverse 3D applications, such as mobile robot navigation, automatic driving \cite{jia2016drivable}, virtual reality \cite{guo2011research} , and augmented reality \cite{oskiper2015augmented}. Stereo image restoration aims to reconstruct high-quality (HQ) stereo images from their low-quality (LQ) counterparts corrupted by various degradation factors. The reconstruction of HQ stereo images can significantly benefit downstream tasks, such as depth estimation, 3D object detection, 3D scene rendering and reconstruction, and image segmentation. Therefore, stereo image restoration has important theoretical significance and practical value.\par

Existing stereo image restoration methods are based on convolutional neural network (CNN) or transformer, which have achieved pronounced effectiveness. However, the two kinds of methods have some drawbacks, respectively. CNN-based methods \cite{jeon2018enhancing, wang2019learning, wang2020parallax, song2020stereoscopic, ying2020stereo, xu2021deep, chen2021cross, wang2021symmetric, yan2020disparity, dai2021feedback} suffer from limited local receptive fields and seem difficult to improve the reconstruction performance of stereo images further. 
Transformer-based methods \cite{yang2022sir, guo2023pft, lin2023steformer} are constrained by the computational burden of intrinsic self-attention layer and restoring both view images. Most transformer-based methods perform cross-view interaction in the same way as PAM in essence or within a local window like SwinIR \cite{liang2021swinir}, limiting transformer’s ability to model long-range context dependencies. \par

In recent years, diffusion models (DMs) \cite{sohl2015deep, ho2020denoising} have aroused great concern due to their superior performance of generating a wide variety of data modalities, including vision \cite{ho2020denoising, song2020denoising, song2020score, dhariwal2021diffusion, esser2023structure}, natural language \cite{touvron2023llama} and audio \cite{huang2023make}. A diffusion probabilistic model is a parameterized Markov chain  trained with variational inference to produce samples matching the data after finite time. In the forward process, Gaussian noise is gradually added to the input image until it becomes pure noise. When the intensity of the Gaussian noise sequence is low enough, the reverse transitions of this chain can be set to conditional Gaussian too. Then the reverse distribution can be parameterized by a simple neural network. Samples can be generated based on the estimated reverse distribution. \par

Motivated by the fact that DMs can model highly complex distributions of natural images, we aim to take advantage of the powerful distribution estimation abilities of DMs to assist the texture recovery in degraded images. However, the utilization of DMs for stereo image restoration is confronted with the following challenges: Firstly, the training and inference of diffusion models are computationally expensive. In stereo image restoration, images of left and right views need to be reconstructed simultaneously, which exacerbates the effect of high computational cost of DM. The majority of DM-based image restoration algorithms \cite{li2022srdiff, saharia2022image, saharia2022palette, dhariwal2021diffusion, niu2023cdpmsr} perform the diffusion process in the image space, which requires thousands of iterations to generate each pixel from noise. It is infeasible to directly apply the paradigm of existing DM-based image restoration algorithms to stereo images. Secondly, existing restoration methods utilize DM to generate the HQ image or its latent embedding, which is then converted to the pixel space. In this manner, DM directly affects the recovery results and is prone to generate some details which are perceptually
high-quality but misaligned with the given LQ image. It can be observed a drop of accuracy related metrics (\textit{e.g.}, PSNR) in these methods compared to CNN and transformer. Thirdly, previous latent compression methods \cite{kingma2013auto, van2017neural, esser2021taming, rombach2022high} are primarily devised for image synthesis, in which the latent representation  focuses on high-level semantics. High-frequency details are removed as redundancy, which are exactly pivotal for low-level restoration. For stereo image restoration, how to preserve the high-frequency details and achieve efficient compression in the meantime is a crucial problem. \par

To address the above issues, we propose DiffStereo, the first diffusion model-based framework for stereo image restoration. Due to the high computational overhead in the image space, we perform the diffusion process in a compressed latent space, whose dimensions are lower than that of the image space. Contrary to previous latent compression methods, DiffStereo is trained to estimate a latent representation preserving high-frequency details of HQ images. The resolution of latent high-frequency representations (LHFR) is kept the same as input images with channel compressed, preventing the inherent texture of input from distortion. The LHFR are then fused into transformer-based restoration network to guide the texture recovery in degraded images, supplementing high-frequency information of corresponding HQ images. The combination of DM and transformer also overcomes the influence of artifacts introduced by DM, ensuring the fidelity of the restored images. \par

In summary, our main contributions are as follows: \par

$\bullet$ We propose a high-frequency aware diffusion model DiffStereo, which is the first DM for stereo image restoration. DiffStereo combines the powerful distribution estimation ability of DM and long-range modeling superiority of transformer and achieves better perceptual quality and higher accuracy. \par

$\bullet$ DiffStereo leverages DM to generate high-frequency representations of HQ images in a compressed latent space, which are fused into transformer-based restoration network. Meanwhile, we design a position encoding scheme to integrate the latent representations into the restoration network. \par

$\bullet$ Extensive experiments conducted on super-resolution (SR), deblurring, and low-light enhancement demonstrate the superior performance of DiffStereo compared with state-of-the-art stereo image restoration methods.

\section{Related Work}
\subsection{Stereo Image Restoration}
The dominant stereo image restoration models are based on CNN or transformers. StereoSR \cite{jeon2018enhancing} is the first deep learning-based algorithm for stereo image SR, which shifts the right view by 1 to 64 pixels and cascades the shifted images with the left view. Then StereoSR fuses the cross-view information by applying convolution to the cascaded images. In this way, for a pixel with parallax less than 64, there must be its corresponding point in one of the shifted images. But pixels with parallax larger than 64 cannot utilize their complementary information, which limits reconstruction performance. PASSR \cite{wang2019learning} integrates epipolar constraints in binocular vision with self-attention mechanism and proposes a parallax attention mechanism (PAM). PAM calculates the similarity between pixels on the same epipolar line and fuses complementary features based on the calculated similarity, which becomes a crucial method for stereo parallax interaction and widely adopted by \cite{wang2020parallax, song2020stereoscopic, ying2020stereo, xu2021deep, chen2021cross, wang2021symmetric}. The other kind of algorithms \cite{yan2020disparity, dai2021feedback} predict pixel-level disparity explicitly and align features of two views according to the disparity map. Recently, transformer has been introduced into stereo image restoration \cite{yang2022sir, guo2023pft, lin2023steformer}. However, there hasn't been any research focusing on diffusion model for stereo image restoration. In this paper, we make the first attempt in this field. \par

\subsection{Diffusion Models}
In recent years, diffusion models have achieved outstanding results in deep generative modeling across various modal domains, such as image \cite{ho2020denoising, song2020denoising, song2020score, dhariwal2021diffusion, saharia2022photorealistic}, video \cite{singer2022make, esser2023structure}, natural language \cite{touvron2023llama}, audio \cite{huang2023make} and 3D \cite{luo2021diffusion}. In image restoration, \cite{li2022srdiff} propose the first DM-based single image SR algorithm SRDiff. SR3 \cite{saharia2022image} adapts DMs to conditional image generation and performs SR through an iterative denoising process. Afterwards, a variety of DM-based image restoration studies have been proposed \cite{saharia2022palette, dhariwal2021diffusion, niu2023cdpmsr, gao2023implicit,}. However, the algorithms listed above perform the diffusion process on whole images, which requires thousands of iteration steps to generate each pixel from noise. LDM \cite{rombach2022high} proposes to perform DM in latent space to improve the restoration efficiency. Furthermore, DiffIR \cite{xia2023diffir} proposes to adopt DM to generate a compact vector for image restoration.

% \subsection{Latent Compression}
% In image synthesis, variational autoencoder (VAE) frameworks \cite{kingma2013auto, van2017neural, esser2021taming} generally learn compressed representations of the input data, whose dimensions are much lower than those of the inputs. Specifically, images are downsampled in the spatial dimension and expanded in the channel dimension. The learned latent space preserves the high-level semantic information but removes high-frequency details, which is exactly pivotal for low-level restoration. Moreover, DiffIR \cite{xia2023diffir} adopts DM to generate a 1D vector, which is efficient but limited in improving the accuracy of reconstruction. 

\section{Method}
\subsection{Overview}
\begin{figure*}[t]\centering
    \includegraphics[width=\textwidth]{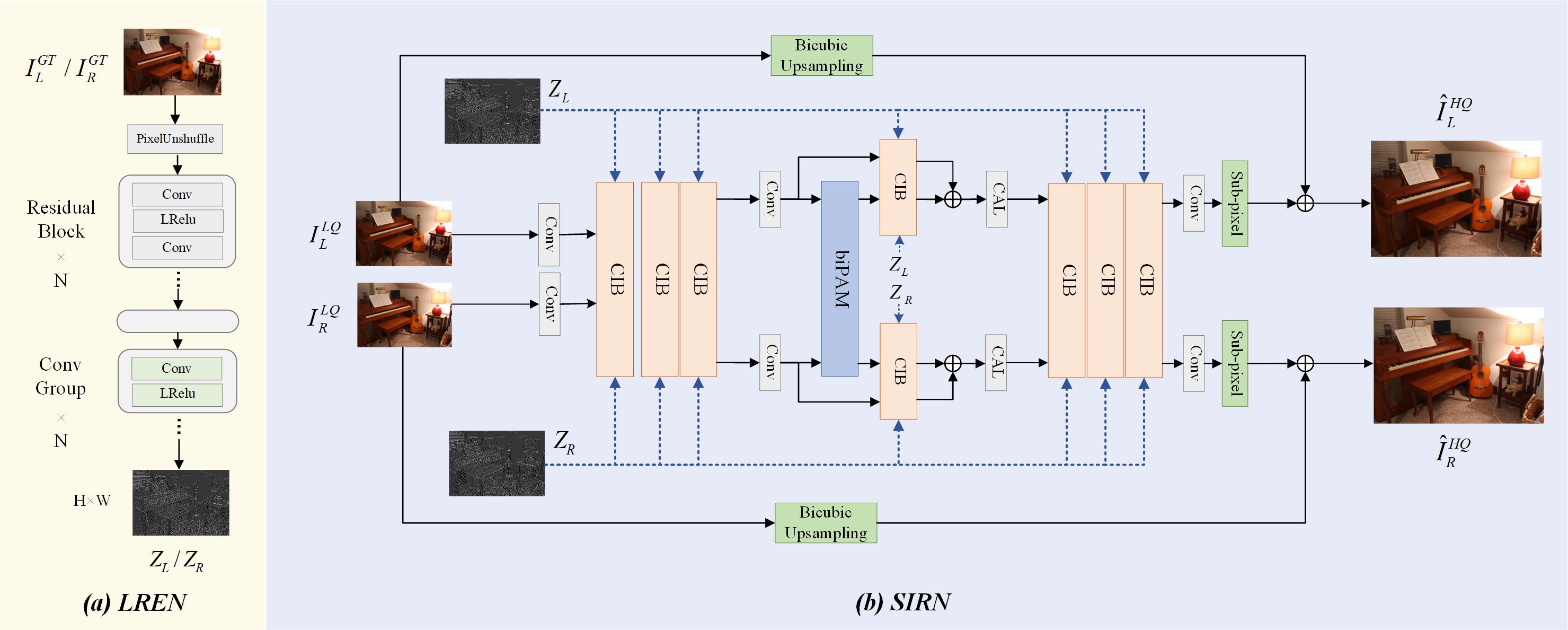}
    \caption{An overview of DiffStereo in training stage one. The latent representation extraction network (LREN) learns a compressed latent space which preserves high-frequency details like structural information and texture in HR stereo images and obtains LHFR of left and right views. The LHFR are then fused into stereo image restoration network (SIRN) and assist the texture recovery.}
    \label{FIG1}
\end{figure*}
Existing DM-based image restoration algorithms follow the paradigm of DM in image synthesis, which generates the whole image or its latent embedding directly. Generating the whole image using a DM typically requires thousands of iterations. In image restoration, it is a waste of computing resources since a part of pixels are already given in the LQ image. Alternatively, if we apply the diffusion process to the embedding of an image, elaborate encoder and decoder are required to convert images between the image space and latent space. Furthermore, the latent compression in image synthesis focuses on high-level semantics with low-level high-frequency details discarded by commonly used downsampling, which isn't suitable for image restoration. \par

The key of DiffStereo is to apply the diffusion model in latent space and estimate latent high-frequency representations (LHFR) of HQ images, which acts as a guidance for the transformer-based restoration network. In this way, DiffStereo combines the strengths of DM and transformer. The DM learns the complex target distribution of stereo images, providing beneficial prior about real distribution. And the regression-based transformer achieves high precision image recovery. \par

Our DiffStereo is composed of three parts: latent representation extraction network (LREN), high-frequency aware DM and stereo image restoration network (SIRN). During training, the LREN first learns high-frequency representations from the HQ stereo images. The diffusion model then learns to estimate the LHFR given LQ images, approximating the output of the LREN. To make the overall framework easier to optimize, we train DiffStereo with a two-stage training strategy, which we will explain in detail in this section. \par

\begin{figure*}[t]\centering
    \includegraphics[width=0.9\textwidth]{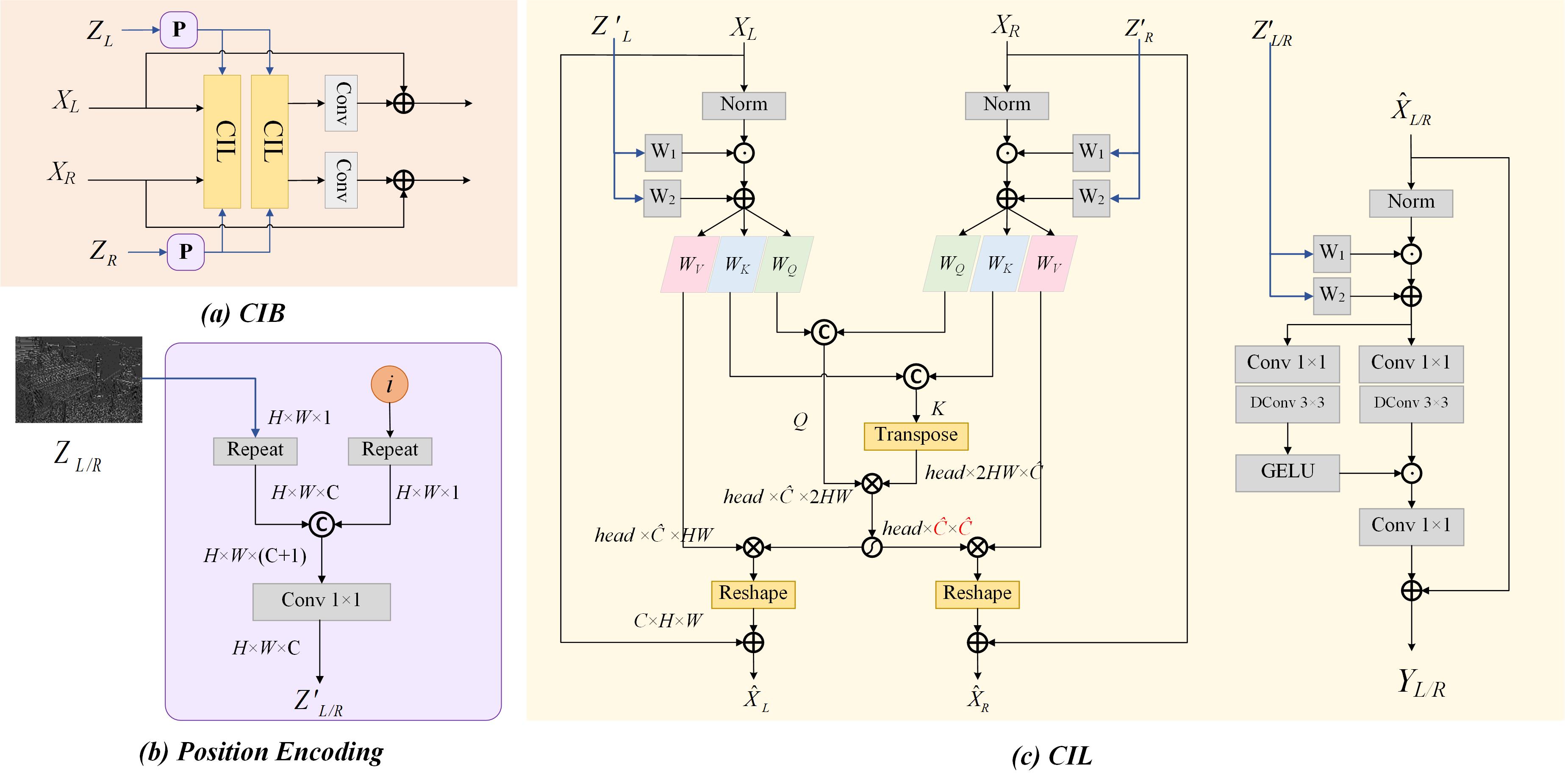}
    \caption{The architecture of our proposed: (a) channel interaction block (CIB), (b) Position Encoding scheme, (c) channel interaction layer (CIL) in CIB.}
    \label{FIG2}
\end{figure*}

\subsection{Training Stage One}
In the first stage, we train the latent representation extraction network (LREN) and stereo image restoration network (SIRN) together to enforce the LREN to learn high-frequency representations conducive to restoration.

\subsubsection{Latent Representation Extraction Network (LREN)}
To extract the LHFR from HQ images, we first downsample the HQ stereo images by PixelUnshuffle operation and extract deep features by stacked residual blocks. Next, several groups of convolution and LeakyReLU are set to compress the channels of the feature maps. The structure of LREN is shown in Fig. \ref{FIG1}(a). The outputs are 2D latent high-frequency representations $Z_L, Z_R \in \mathbb{R}^{H\times W}$, acting as guidance for stereo image restoration. 
% \begin{equation}
%     Z_L = LREN(PixelUnshuffle(I_L^{GT}))
% \end{equation}
% \begin{equation}
%     Z_R = LREN(PixelUnshuffle(I_R^{GT}))
% \end{equation}
\par

\subsubsection{Stereo Image Restoration Network (SIRN)}
For stereo image restoration, we design a transformer-based network considering the fact that transformer can model long-range pixel dependencies. Restormer \cite{zamir2022restormer} introduces a multi-Dconv head transposed attention with linear complexity which applies self-attention across feature dimension rather than the spatial dimension. Inspired by the idea, we devise a channel interaction block (CIB) as the basic unit of our SIRN. CIB concatenates the features of the two views and uses all pixels from the two views to calculate a channel attention map shared by the two views. In the process of stereo imaging, since the left and right images are shot, transmitted and stored at the same time, they suffer from the same degradation and have the same distribution. Thus the channel covariance should be the same too. Taking account of all pixels in the two views is beneficial to obtaining more accurate channel covariance. \par

In each CIB, layer normalization is first applied to the input features $X_L$ and $X_R \in \mathbb{R}^{C \times H \times W}$. 1×1 convolutions encode pixel-wise channel context, followed by 3×3 depth-wise convolutions aggregating channel-wise spatial context. We reshape output of the convolutions to $\mathbb{R}^{head \times \hat{C}  \times HW}$ and obtain query (Q), key (K), and value (V) of the two views. Then we concatenate $Q_L$ and $Q_R$, $K_L$ and $K_R$ along the spatial dimension,
\begin{equation} 
\begin{aligned}
    Q &= \operatorname{Concat}(Q_L, Q_R) \\
      &= \operatorname{Concat}(W^Q_d W^Q_p \text{LN}(X_L), W^Q_d W^Q_p \text{LN}(X_R))
\end{aligned}
\end{equation}

\begin{equation}
\begin{aligned}
    K &= \operatorname{Concat}(K_L, K_R) \\
      &= \operatorname{Concat}(W^K_d W^K_p \text{LN}(X_L), W^K_d W^K_p \text{LN}(X_R))
\end{aligned}
\end{equation}

\begin{equation}
    V_{L/R} = W^V_d W^V_p \text{LN}(X_{L/R})
\end{equation}
where $W_{p}^{(\cdot )}$  is 1×1 point-wise convolution and $W_{d}^{(\cdot )}$ is 3×3 depth-wise convolution. Then multi-Dconv head transposed attention is calculated and generates channel attention map $A \in \mathbb{R}^{head \times \hat{C} \times \hat{C}}$ shared by the two views and enhanced stereo features $\hat{Y}_L$ and $\hat{Y}_R \in \mathbb{R}^{C \times H \times W}$,
\begin{equation}
    \hat{Y}_{L/R} = \text{Attention}(Q,K,V_{L/R}) 
    = \text{Softmax}(Q \cdot K^T /w) \cdot V_{L/R}
\end{equation}
where $w$ is a learnable scaling factor. $\hat{Y}_L$ and $\hat{Y}_R$ are then sent into gated-Dconv feed-forward network (GDFN) \cite{zamir2022restormer} and generate the final output of CIB $Y_L$ and $Y_R \in \mathbb{R}^{C \times H \times W}$ , which is defined as,
\begin{equation}
    Y_{L/R} = \text{GDFN} (\hat{Y}_{L/R} + X_{L/R})
\end{equation}

\subsubsection{Position Encoding Scheme.}
We propose a position encoding scheme when integrating the LHFR into restoration network in order to provide distinctive priors for CIBs of different depths. The position index $i$ is generated based on the depth of each CIB in SIRN. The CIB of the first layer corresponds to index $0$ and the index increases in ascending order in following layers. We copy the index into a matrix $I \in \mathbb{R}^{H \times W \times 1}$. Before integrating the LHFR into each CIB, we extend the dimension of $Z_L, Z_R$ to $\mathbb{R}^{H\times W \times 1}$ and copy them $C$ times along the channel and concatenate the repeated LHFR with $I$. The two features are fused by 1×1 point-wise convolution and generate encoded $Z'_L, Z'_R \in \mathbb{R}^{H\times W \times C}$, the same shape as $X_L$ and $X_R$. Then $Z'_L, Z'_R \in \mathbb{R}^{H\times W \times C}$ are fused with the features in CIB in a way of spatial modulation, 
\begin{equation}
    X'_{L/R} = W_1 Z'_{L/R} \odot \text{LN}(X_{L/R}) + W_2 Z'_{L/R}
\end{equation}
where $W_1$ and $W_2$ denote weights of convolution layers.

\begin{figure*}[t]\centering
    \includegraphics[width=0.9\textwidth]{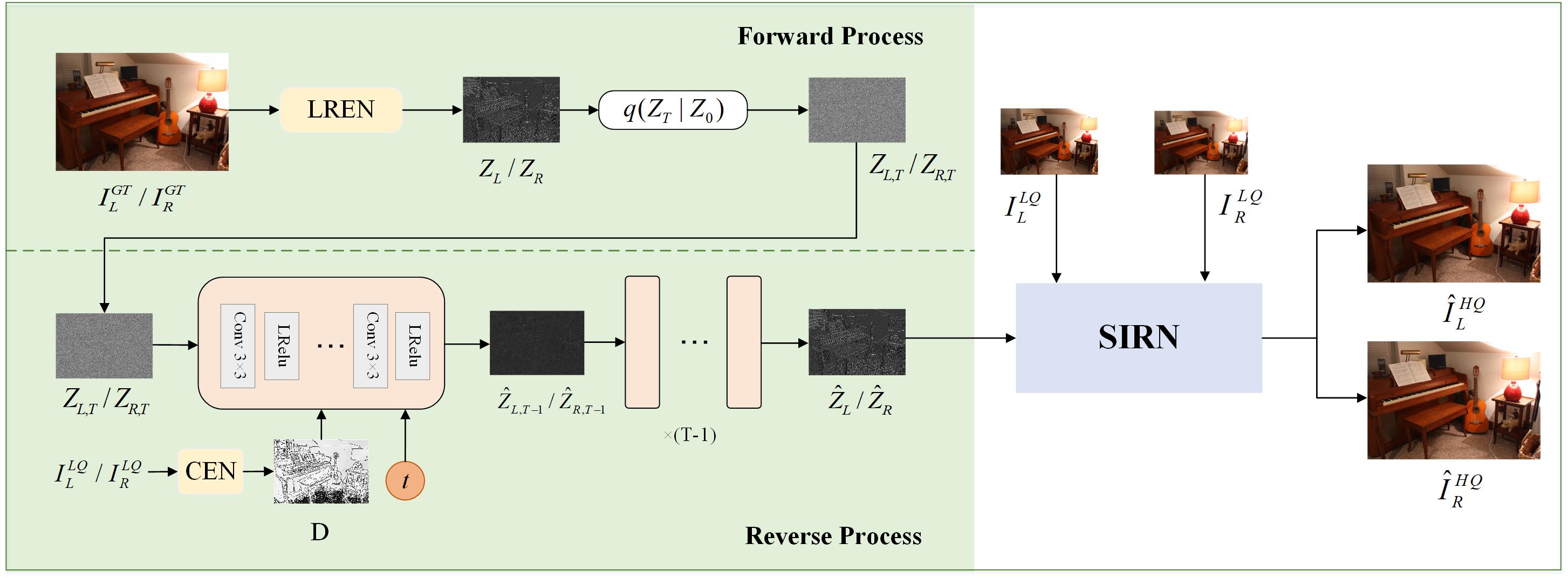}
    \caption{An overview of DiffStereo in training stage two. The DM learns to estimate the LHFR extracted by pretrained LREN, whose parameters are frozen in stage two. During inference, the DM estimates LHFR from pure Gaussian noise under the guidance of LR stereo images.}
    \label{FIG3}
\end{figure*}
\subsection{Training Stage Two}
In the second stage, the high-frequency aware DM and stereo image restoration network (SIRN) are optimized together to enforce DM to estimate the high-frequency latent representations given LQ images, approximating the output of the LREN. \par

Using the pretrained LREN in Stage One, the groundtruth LHFR are extracted and denoted as $\bm{Z}_L, \bm{Z}_R \in \mathbb{R}^{H\times W}$. We apply the forward diffusion process on $\bm{Z}_L, \bm{Z}_R$ and the noised LHFR are sampled by,
\begin{equation}
    q(\bm{Z}_{L,T}| \bm{Z}_L) = \mathcal{N}(\bm{Z}_{L,T}; \sqrt{\bar{\alpha}_T} \bm{Z}_L, (1-\bar{\alpha}_T) \bm{I})
\end{equation}

\begin{equation}
    q(\bm{Z}_{R,T}| \bm{Z}_R) = \mathcal{N}(\bm{Z}_{R,T}; \sqrt{\bar{\alpha}_T} \bm{Z}_R, (1-\bar{\alpha}_T) \bm{I})
\end{equation}
where $T$ is the total number of iterations, $\bm{Z}_{L,T}$ and $\bm{Z}_{R,T}$ are the LHFR of left and right view at time step $T$. $\bar{\alpha}_T = {\textstyle \prod_{i=0}^{T}} \alpha_i$, $\alpha_i$ is the variance schedule. \par

In the reverse process, the diffusion model learns the reverse transitions of the Markov chain, which is a conditional Gaussian distribution denoted as,
\begin{equation}
    q(\bm{Z}_{t-1}|\bm{Z}_{t}, \bm{Z}_0) = \mathcal{N} (\bm{Z}_{t-1}; \mathbf{\mu}_t(\bm{Z}_t, \bm{Z}_0), \sigma^2_t \bm{I}),
\end{equation}

\begin{equation}
\mathbf{\mu}_t(\bm{Z}_t, \bm{Z}_0) = \frac{1}{\sqrt{\alpha_t}}(\bm{Z}_t - \frac{1-\alpha_t}{\sqrt{1-\bar{\alpha}}_t} \epsilon_t), \sigma^2_t = \frac{1-\bar{\alpha}_{t-1}}{1-\bar{\alpha}_t} \beta_t.
\end{equation}
where $\epsilon_t$ represents the noise added in $\bm{Z}_t$, which is the only unknown term in the posterior distribution. The subscripts $L$ and $R$ denoting left and right view are omitted for simplicity. We can use a particularly simple neural network $\theta$ to estimate $\hat{\epsilon}_t$ and $\bm{Z}_{t-1}$ can be sampled by parameterizing the Gaussian distribution, 
\begin{equation}
  \hat{\bm{Z}}_{t-1} = \frac{1}{\sqrt{\alpha_t} }(\hat{\bm{Z}}_t- \frac{1-\alpha_t}{\sqrt{1-\bar{\alpha}_t } }  \hat{\epsilon}_t ) 
\label{eqreverse}
\end{equation}
The process in Eq. (\ref{eqreverse}) is iterated $T$ times until the pure LHFR $\hat{\bm{Z}}_L, \hat{\bm{Z}}_R$ are recovered. It is worth noting that the random term following standard normal distribution with a coefficient of $\sigma_t$ is deleted for better image fidelity. \par

The denosing network $\theta$ is expected to predict a sequence of noise to recover a LHFR corresponding to the given LQ image. Thus we first adopt a simple CNN network, denoted as CEN (condition extraction network) to obtain a conditional feature $D \in \mathbb{R}^{H \times W}$ from given LQ image. $D$ is then sent into the denoising network $\theta$ for noise prediction with the encoded spatial information of LQ image as well as the timestep $t$, which is described as,
\begin{equation}
    \hat{\epsilon}_t = \theta(Concat(\hat{\bm{Z}}_t, D, t))
\end{equation}

\begin{table*}[t] \centering 
\caption{Quantitative comparisons for ×4 SR with PSNR/SSIM/FID values. Higher PSNR/SSIM and lower FID mean better performance. \textcolor{red}{Red} and \textcolor{blue}{blue} colors represent the best and second best performance, respectively.}
\label{tab1}
\begin{tabular}{ccccc}
\toprule
\multirow{2}{*}{Method} & \multicolumn{4}{c}{PSNR $\uparrow$ / SSIM $\uparrow$ / FID $\downarrow$ / LPIPS $\downarrow$}  \\ \cline{2-5} 
& Flickr1024   & KITTI2012    & KITTI2015    & Middlebury   \\ \hline
EDSR   
& 23.46/0.7285/0.1797/0.2672 
& 26.35/0.8015/0.2655/0.1965 
& 26.04/0.8039/0.1959/0.1565
& 29.23/0.8397/0.1985/0.1584 \\
RDN    
& 23.47/0.7295/0.1529/\textcolor{blue}{0.2648}
& 26.32/0.8014/\textcolor{blue}{0.2621}/\textcolor{blue}{0.1949}
& 26.04/0.8043/0.1990/\textcolor{blue}{0.1547}
& 29.27/0.8404/0.1926/\textcolor{blue}{0.1577} \\
RCAN    
& 23.48/0.7286/0.1828/0.2705 
& 26.44/0.8029/0.2773/0.2042 
& 26.22/0.8068/0.2034/0.1590 
& 29.30/0.8397/0.1994/0.1603 \\
StereoSR   
& 21.70/0.6460/0.1856/0.2937 
& 24.53/0.7556/0.3049/0.2072 
& 24.21/0.7511/0.2554/0.1931 
& 27.64/0.8022/0.2496/0.1709 \\
PASSRnet  
& 23.31/0.7195/0.1638/0.2871 
& 26.34/0.7981/0.2622/0.2095 
& 26.08/0.8002/\textcolor{red}{0.1880}/0.1606 
& 28.72/0.8236/0.2305/0.1969 \\
SRRes+SAM 
& 23.27/0.7233/\textcolor{red}{0.1450}/0.2772 
& 26.44/0.8018/0.2665/0.2122 
& 26.22/0.8054/0.1962/0.1667 
& 28.83/0.8290/\textcolor{blue}{0.1966}/0.1721 \\
iPASSR   
& 23.44/0.7287/0.1815/0.2713 
& 26.56/0.8053/0.2674/0.2076 
& 26.32/0.8084/0.1971/0.1614 
& 29.16/0.8367/0.2048/0.1657 \\
SSRDE-FNet     
& \textcolor{blue}{23.55/0.7346}/0.1765/0.2695 
& \textcolor{red}{26.69/0.8091}/0.2749/0.2074
& \textcolor{red}{26.46}/\textcolor{blue}{0.8133}/0.2000/0.1566 
& \textcolor{blue}{29.34/0.8411}/0.2039/0.1713 \\
SIR-Former  
& 23.52/0.7305/\ \ \ \ - \ \ \ \ /\ \ \ \ - \ \ \ \ 
& \textcolor{blue}{26.68}/0.8077/\ \ \ \ - \ \ \ \ /\ \ \ \ - \ \ \ \ 
& 26.42/0.8098/\ \ \ \ - \ \ \ \ /\ \ \ \ - \ \ \ \ 
& 29.32/0.8407/\ \ \ \ - \ \ \ \ /\ \ \ \ - \ \ \ \    \\
DiffStereo (Ours) 
& \textcolor{red}{23.60}/\textcolor{red}{0.7379}/\textcolor{blue}{0.1460}/\textcolor{red}{0.2603} 
& 26.67/\textcolor{blue}{0.8081}/\textcolor{red}{0.2612}/\textcolor{red}{0.1944 }
& \textcolor{red}{26.46}/\textcolor{red}{0.8148}/\textcolor{blue}{0.1961}/\textcolor{red}{0.1516}
& \textcolor{red}{29.41}/\textcolor{red}{0.8441}/\textcolor{red}{0.1913}/\textcolor{red}{0.1559}  \\ 
\bottomrule
\end{tabular}
\end{table*}

\subsection{Loss Functions}
\textbf{Reconstruction Loss.} The reconstruction loss is the $L_1$ distance between the restored and groundtruth stereo images:
\begin{equation}
    \mathcal{L}_{rec} = \left \| \hat{I}_{L}^{HQ} - I_{L}^{GT} \right \|_1 + \left \| \hat{I}_{R}^{HQ} - I_{R}^{GT} \right \|_1 
\end{equation}
where $\hat{I}_{L}^{HQ}$ and $\hat{I}_{R}^{HQ}$ are restored high-quality left and right images, $I_{L}^{GT}$ and $I_{R}^{GT}$ are the groundtruth left and right images.
\par

\textbf{Parallax Related Loss.}
Following \cite{wang2021symmetric}, we use smoothness loss, residual photometric loss, residual cycle loss and residual stereo consistency loss to supervise the network to estimate disparity correlations between left and right view accurately. We denote these terms as parallax related loss $\mathcal{L}_{para}$. 

\textbf{Diffusion Related Loss.}
The optimization of denoising diffusion probabilistic models in \cite{ho2020denoising} is implemented by,
\begin{equation}
    \nabla_{\theta} \left \| \epsilon - \epsilon_{\theta}(\sqrt{\bar{\alpha}_t} x_0 + \sqrt{1-\bar{\alpha}_t} \epsilon, t ) \right \| ^2
\end{equation}
where $t$ is uniformly sampled from $\{1,...,T\}$. Since in DiffStereo, the DM only needs to estimate LHFR whose dimensions are much lower than that of the original image, it's feasible to run all iterations of DM at a time and train the DM by the distance between estimated $\hat{\bm{Z}}_L, \hat{\bm{Z}}_R$ by DM and $\bm{Z}_L, \bm{Z}_R$ extracted by LREN,
\begin{equation}
    \mathcal{L}_{diff} = 
    \| \hat{\bm{Z}}_L - \bm{Z}_L\|_{1} + \| \hat{\bm{Z}}_R - \bm{Z}_R \|_{1}
\end{equation} 
\par

In the first stage, reconstruction loss and parallax related loss are utilized to train the LREN and SIRN, which is defined as follows:
\begin{equation}
    \mathcal{L}_{stage1} = \mathcal{L}_{rec} + \lambda_1 * \mathcal{L}_{para} 
\end{equation}
where $\lambda_1$ represents the weight of parallax related loss is set to 0.1 in this work. In the second stage, diffusion related loss is also included in the overall training objective, 
\begin{equation}
    \mathcal{L}_{stage2} = \mathcal{L}_{rec} + \lambda_1 * \mathcal{L}_{para} + \lambda_2 * \mathcal{L}_{diff}
\end{equation}
where $\lambda_2$ represent the weight of diffusion model related loss, set to 0.25 in this work.

\begin{figure*}[t]\centering
    \includegraphics[width=1\textwidth]{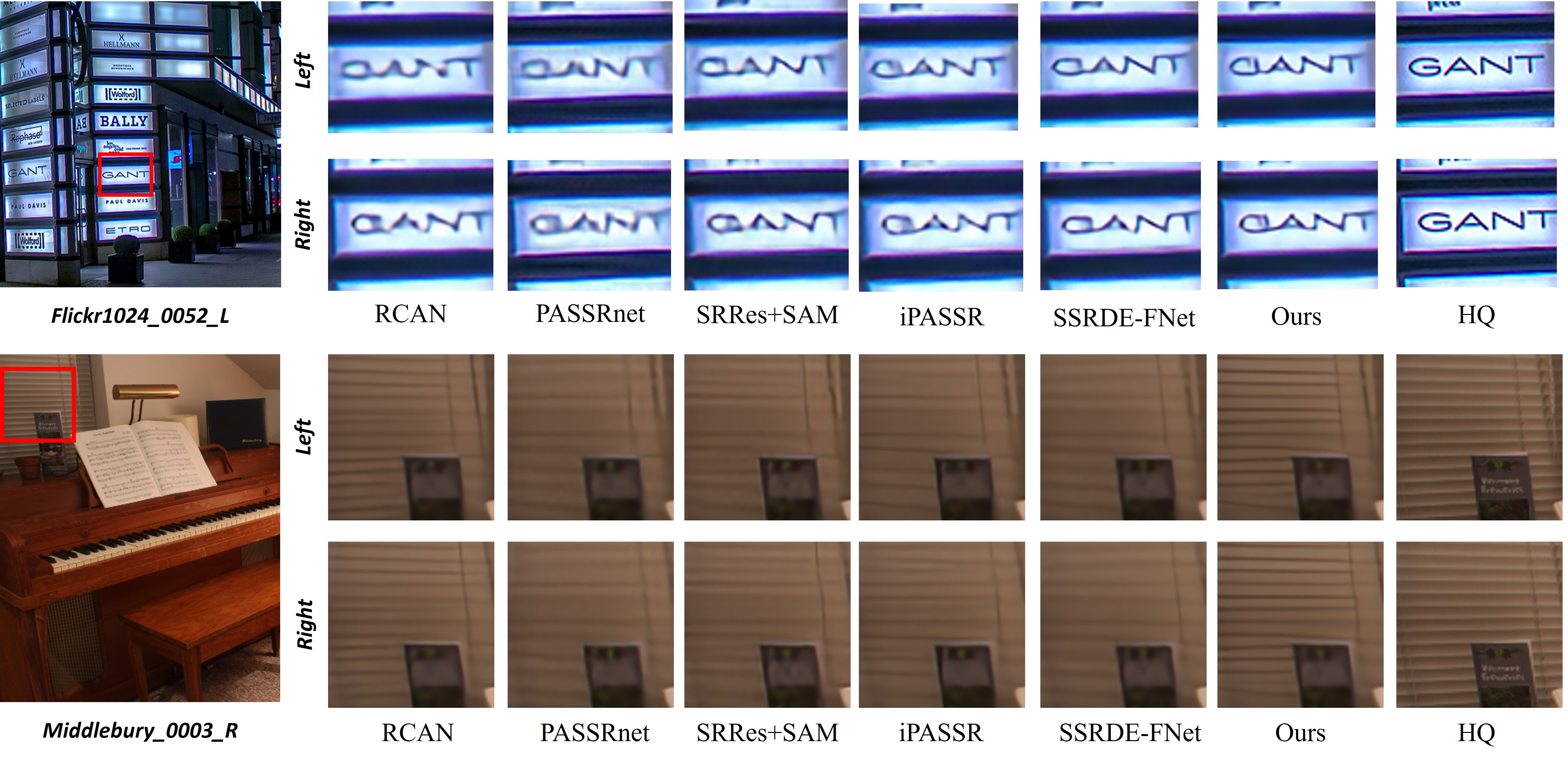}
    \caption{Visiual comparisons for ×4 SR by different methods on Flickr1024 and Middlebury dataset.}
    \label{FIG4}
\end{figure*}

\begin{table*}[t]
\caption{Quantitative comparisons for deblurring with PSNR/SSIM values. Higher PSNR and SSIM mean better performance. \textcolor{red}{Red} color represents the best performance, respectively.}
\label{tab2}
\begin{tabular}{cccccccccc}
\toprule
\multirow{2}{*}{Method} & \multirow{2}{*}{Params} & \multicolumn{3}{c}{Left}  &  & \multicolumn{4}{c}{(Left + Right) / 2}          \\ \cline{3-5} \cline{7-10} 
&                         & KITTI2012 & KITTI2015 & Middlebury &  & Flickr1024 & KITTI2012 & KITTI2015 & Middlebury \\ \hline
PASSRnet   & 1.37M & 37.87/0.9675 &  36.94/0.9699  &    40.33/0.9768         &  &      37.10/0.9751      &    37.88/0.9682     &    37.53/0.9726    &  40.34/0.9768     \\
iPASSR   &  1.43M  &   38.63/0.9719  &  38.07/0.9766   &    41.13/0.9802  &  &    38.09/\textcolor{red}{0.9793}     &  38.60/0.9723    &  38.71/0.9788      &   41.17/0.9802  \\
DiffStereo (Ours)  &  2.16M    
& \textcolor{red}{38.78/0.9721}        
& \textcolor{red}{38.33/0.9771}    
& \textcolor{red}{41.34/0.9804}         &  
& \textcolor{red}{38.13}/0.9790         
& \textcolor{red}{38.74/0.9724}      
& \textcolor{red}{38.98/0.9793}       
& \textcolor{red}{41.37/0.9804}  \\     
\bottomrule
\end{tabular}
\end{table*}

\begin{figure*}[t]\centering
    \includegraphics[width=1\textwidth]{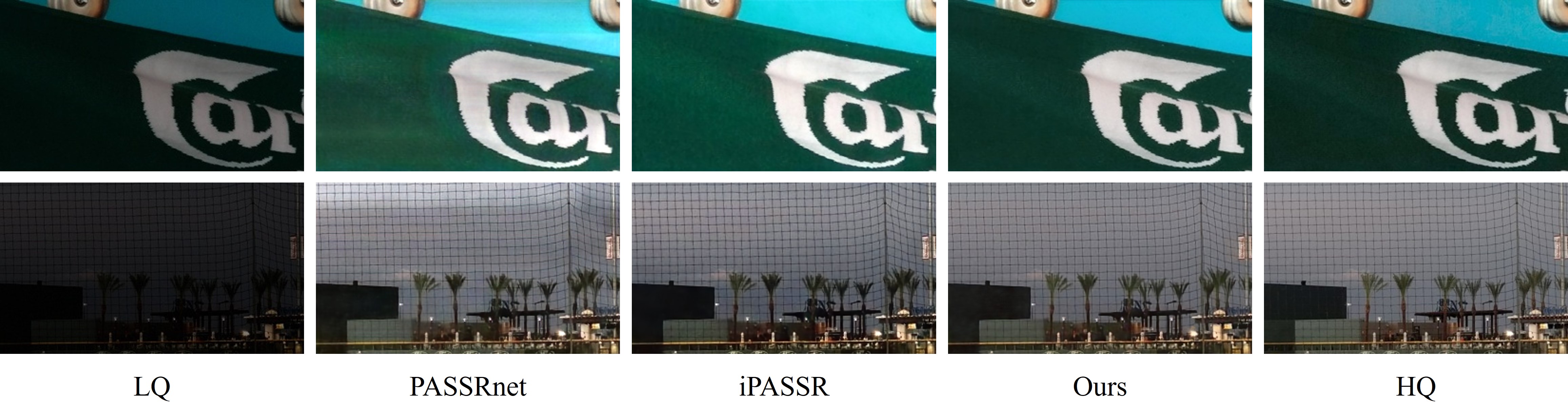}
    \caption{Visiual comparisons in Low-Light Enhancement by different methods.}
    \label{FIG5}
\end{figure*}

\section{Experiments}
\subsection{Datasets and Implementation Details}
\textbf{Datasets.} For SR, we construct our training and test set following the previous works \cite{jeon2018enhancing, wang2019learning, ying2020stereo, wang2021symmetric}. 800 images from the training set of Flickr1024 \cite{wang2019flickr1024} and 60 images from Middlebury \cite{scharstein2014high} make up the training data. For test set, we use 5 images from
Middlebury \cite{scharstein2014high}, 20 images from KITTI 2012 \cite{geiger2012we}, 20 images from KITTI 2015 \cite{menze2015object} and all the 112 images from the test
set of Flickr1024 \cite{wang2019flickr1024}. Low-resolution images are generated by bicubic downsampling the HQ images. For deblurring, we use the same HQ images as in SR. And we convolve the HQ images with a Gaussian blur kernel whose kernel size is 15×15 and $\sigma$ is 1.0 to generate LQ images. For low-light enhancement, we use the synthetic Holopix50k \cite{huang2022low} dataset, which is generated by adopting linear transformation and gamma correction for the normal-light images \cite{hua2020holopix50k} to reduce lightness and then adding Gaussian-Poisson mixed noise. For all of the three tasks, we crop LQ images into 30×90 patches with a stride of 20 before training. Randomly flipping horizontally and vertically is applied for data augmentation.\par

\begin{table}[t]  \centering
\caption{Comparisons of different low-light enhancement methods in terms of PSNR, SSIM, FID and LPIPS.}
\label{Tab3}
\begin{tabular}{c|ccc}
\toprule
Method & PASSRnet & iPASSR & DiffStereo (Ours) \\ \hline
PSNR $\uparrow$ & 25.43 & 26.69 & \textcolor{red}{27.48} \\
SSIM $\uparrow$ & 0.8864 & 0.9048 & \textcolor{red}{0.9076} \\ 
FID $\downarrow$ & 0.0910 &  0.0586 & \textcolor{red}{0.0454} \\
LPIPS $\downarrow$ & 0.0557  & \textcolor{red}{0.0446} & 0.0459 \\
\bottomrule
\end{tabular}
\end{table}

\textbf{Implementation Details.} DiffStereo is implemented in PyTorch and trained with two NVIDIA RTX 3090 GPU. We use the Adam optimizer with $\beta_1 = 0.9$ and $\beta_2 = 0.999$ and the batch size is set to 48. DiffStereo is trained for 90 epochs in the first stage and 300 epochs in the second stage.

\textbf{Evaluation Metric.} Following previous works, we adopt peak signal-to-noise ratio (PSNR) and structural similarity index metrics (SSIM) to measure the accuracy of reconstructed images. Furthermore, we use Frechet inception distance score (FID) to measure the distance between the real image distribution and the restored image distribution. Learned perceptual image patch similarity (LPIPS) is also adopted to evaluate the perceptual quality of reconstructed images.

\subsection{Comparisons with State-of-the-Art Methods}
In this section, we compare the proposed DiffStereo with SOTA restoration methods on SR, deblurring and low-light enhancement. For fair comparisons, all the methods below are trained with the same datasets and data augumentation strategies. 

\subsubsection{Stereo Image Super-Resolution}
As displayed in Table \ref{tab1}, we compare the proposed DiffStereo with single image SR methods EDSR \cite{lim2017enhanced}, RDN \cite{zhang2018residual}, RCAN \cite{zhang2018image} and stereo image SR methods StereoSR \cite{jeon2018enhancing}, PASSRnet \cite{wang2019learning}, SRRes+SAM \cite{ying2020stereo}, iPASSR \cite{wang2021symmetric}, SSRDE-FNet \cite{dai2021feedback} and SIR-Former \cite{yang2022sir}. FID and LPIPS of SIR-Former aren't given because its source code isn't publicly available. In terms of reconstruction accuracy, DiffStereo achieves comparable PSNR/SSIM on KITTI2012 and KITTI2015 datasets and better PSNR/SSIM on Flickr1024 and Middlebury datasets. At the aspect of FID, DiffStereo surpasses SOTA method SSRDE-FNet by 0.0305, 0.0137, 0.0039, 0.0126, which means the stereo image distribution estimated by DiffStereo is closer to the real distribution than that of SSRDE-FNet. This is owing to the powerful distribution estimation ability of diffusion model. For LPIPS, DiffStereo outperforms other methods on all the four datasets, fully proving that the introduction of DM can benefit the perceptual quality of restored images. In general, our DiffStereo achieves the best balance between reconstruction accuracy, distribution estimation and perceptual quality by the effective combination of transformer and DM. Qualitative comparisons are shown in Fig. \ref{FIG4}. It's obvious that our DiffStereo can recover richer details and clearer edges compared with other SOTA methods. \par

\subsubsection{Stereo Image Deblurring}
To evaluate the generalizaiton ability of DiffStereo, we also compare it on stereo deblurring task with representative methods PASSRnet \cite{wang2019learning} and iPASSR \cite{wang2021symmetric}. It can be observed from Table \ref{tab2} that DiffStereo surpasses iPASSR by up to 0.27 dB, demonstrating the superiority of DiffStereo in deblurring.

\begin{figure*}[t]\centering
    \includegraphics[width=1\textwidth]{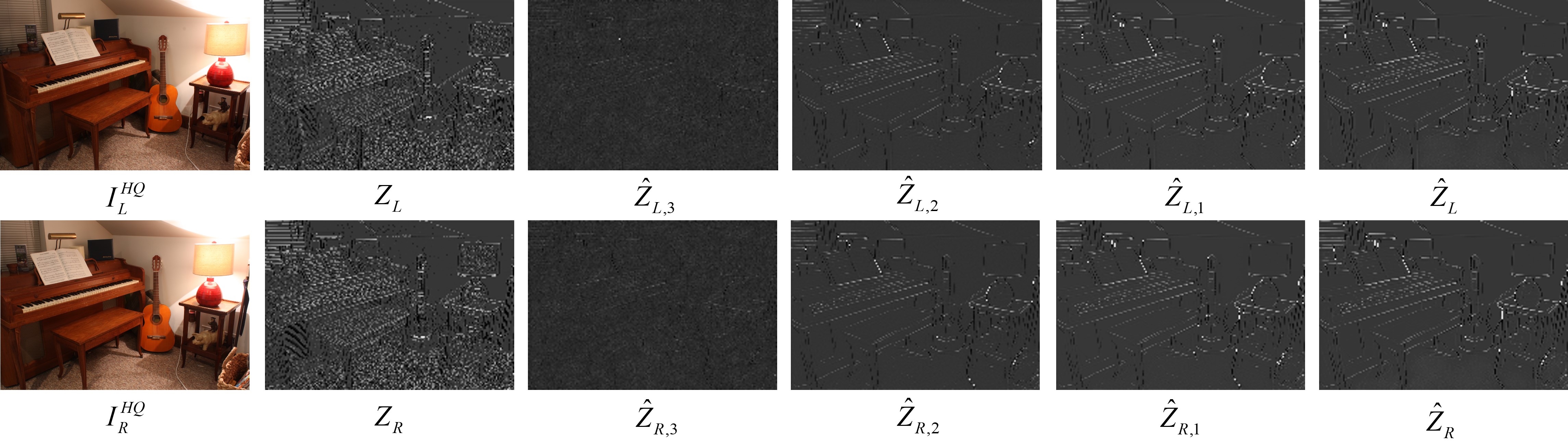}
    \caption{Visualizations of LHFR in different training stages and different timesteps of denoising. The total timestep $T$ is set to 4 since more interations do not provide further improvement.}
    \label{FIG6}
\end{figure*}

\begin{table*}[t]\centering 
\caption{Quantitative comparisons for different latent representations and Position Encoding ($i.e., $ PE) in terms of PSNR/SSIM.}
\label{Tab4}
\begin{tabular}{cccccccccc}
\toprule
& 1D vector & LHFR & PE & Stage & Flickr1024 & KITTI2012 & KITTI2015 & Middlebury \\ \hline
1 & $\times$ & $\times$  & $\times$ & - & 23.53/0.7347 & 26.64/0.8079 & 26.39/0.8117 & 29.34/0.8409 \\
2 & $\checked$ & $\times$  & $\times$ & 1 & 23.59/0.7366 & 26.70/0.8083 & 26.51/0.8135 & 29.33/0.8411 \\
3 & $\times$ & $\checked$  & $\times$ & 1 & 26.98/0.8704 & 29.43/0.8883  & 28.71/0.8860 & 32.85/0.9227 \\
4 & $\times$ & $\checked $ & $\checked$ & 1 & 27.15/0.8748  & 29.53/0.8902 & 28.78/0.8878 & 33.07/0.9259 \\ 
\bottomrule
\end{tabular}
\end{table*}

\subsubsection{Stereo Image Low-Light Enhancement}
On stereo low-light enhancement task, DiffStereo achieves a significant improvement of 0.79 dB in terms of PSNR compared with iPASSR. The obvious gain of PSNR, SSIM and FID substantiates the generalizaiton of our DiffStereo. And DiffStereo is able to adapt to changes in light conditions in a wide variety of realistic scenarios of Holopix50k.

\begin{table}[t]
\caption{Comparisons of DiffStereo with and without ($i.e., $ w/o) LHFR ($i.e., $ LR) on SR, deblurring and low-light enhancement ($i.e., $ LLE).}
\label{Tab5}
\begin{tabular}{c|c|cc|cc}
\toprule
\multirow{2}{*}{Task}  & \multirow{2}{*}{Dataset} & \multicolumn{2}{c|}{PSNR} & \multicolumn{2}{c}{SSIM} \\ \cline{3-6} 
&                   & w/o LR      & with LR     & w/o LR      & with LR      \\ \hline
\multicolumn{1}{c|}{\multirow{4}{*}{SR}} & Flickr1024        &      23.53 &  \textbf{23.60}       &     0.7342        &    \textbf{0.7379}     \\ \cline{2-2}
\multicolumn{1}{c|}{}     & KITTI2012     &  26.65 &  \textbf{26.67}   &   0.8079     &   \textbf{0.8081}    \\ \cline{2-2}
\multicolumn{1}{c|}{}      & KITTI2015    &  26.41 &  \textbf{26.46}       &      0.8118       &      \textbf{0.8148}    \\ \cline{2-2}
\multicolumn{1}{c|}{}      & Middlebury        &          29.30  &    \textbf{29.41}    &        0.8399    &    \textbf{0.8441}       \\ \hline
\multirow{4}{*}{Deblur}     & Flickr1024        &   37.79  &  \textbf{38.13}       &      0.9790       &    \textbf{0.9790}        \\ \cline{2-2}
& KITTI2012         &        38.43       &  \textbf{38.74}     &     \textbf{0.9725}       &          0.9724    \\ \cline{2-2}
& KITTI2015      &    38.42   &  \textbf{38.98}      &    0.9790         &   \textbf{0.9793}           \\ \cline{2-2}
& Middlebury  &  40.80     &    \textbf{41.37}       &     0.9803        &  \textbf{0.9804}         \\ \hline
\multicolumn{1}{c|}{LLE}                 
& Holopix50k    & 27.42  & \textbf{27.48} & 0.9067        &      \textbf{0.9076}     \\ 
\bottomrule
\end{tabular}
\end{table}
\par

\subsection{Ablation Study}
\subsubsection{Effectiveness of Latent High-Frequency Representations (LHFR)}
We conduct the following three experiments to illustrate the importance of high-frequency aware compression and the effectiveness of LHFR: (1) The LHFR estimated by diffusion model are not introduced into SIRN. (2) The LHFR are replaced with 1D channel vectors $Z \in \mathbb{R}^C$. (3) The original LHFR $Z \in \mathbb{R}^{H \times W}$ used in DiffStereo. Comparing (1) with (2) and (3), it is obvious that the latent representations estimated by DM benefit the quality of reconstruction in terms of PSNR and SSIM. Particularly, the LHFR improve the restoration performance significantly. Furthermore, in our settings, $C=256$, $H=30$ and $W=90$, so the compression ratios of the two kinds of latent representations are almost the same during training. But the LHFR work much better than 1D vectors, which means the spatial structure information is more rewarding than global semantics for restoration. As shown in Fig. \ref{FIG6}, we visualize the LHFR $Z_L, Z_R$ extracted from HQ images in the first stage and $\hat{Z}_L, \hat{Z}_R$ estimated by DM at various timesteps. Under the constraint of reconstruction loss, the latent representations learned by DiffStereo indeed focus on high-frequency edges. And given LR images, DM is capable of perceiving high-frequency information in HQ images, taking only a few denoising iterations. \par

Furthermore, we assess the role of LHFR generated by DM on SR, deblurring and low-light enhancement. The quantitative results are shown in Table \ref{Tab5}. On all datasets for all tasks, there is a consistent and significant improvement of PSNR. Especially, the LHFR bring a gain of 0.57 dB on Middlebury dataset for deblurring.

\subsubsection{Effectiveness of the Position Encoding}
To explore the effectivity of position encoding proposed in DiffStereo, we evaluate PSNR and SSIM with and without the position encoding scheme. As displayed in Table \ref{Tab4}, the position encoding scheme can further improve the restoration performance on the basis of LHFR comparing (3) and (4). It is suggested that providing distinctive priors in different depths of image restoration network can help the network to recover image details better.

\subsubsection{Effectiveness of CIB in Stereo Image Restoration Network}
\begin{table}[t]
\caption{Comparisons of different structures of SIRN.}
\label{Tab6}
\begin{tabular}{c|cccc}
\toprule
 & Flickr1024 & KITTI2012 & KITTI2015 & Middlebury \\ \hline
NAFB & 23.24 & 26.36 &  26.05 & 29.05  \\
RDB & 23.43 & 26.56 & 26.30 & 29.18  \\
CIB & \textbf{23.59} & \textbf{26.64} & \textbf{26.41} & \textbf{29.39} \\ \hline
\end{tabular}
\end{table}
In order to study which type of restoration network matches DM better, we replace the proposed CIB with commonly used residual dense block (RDB) and nonlinear activation free block (NAFB) \cite{chen2022simple} in image restoration. It can be observed from Table \ref{Tab6} that CIB outperforms both RDB and NAFB by a significant margin, highlighting the effectiveness of our proposed CIB. The results demonstrate that the latent representations generated by DM is more suitable to a transformer-based structure. The combination of the CIB block and DM can achieve superior performance in image restoration tasks.

\section{Conclusion}
In this paper, we propose a novel high-frequency aware diffusion model for stereo image restoration as the first attempt of DM in this domain. Contrary to previous latent compression methods focusing on high-level semantics, DiffStereo is trained to preserve high-frequency details of HQ images and utilizes the DM to generate latent high-frequency representations compressed from groundtruth HQ images, which are sent into a transformer-based stereo image restoration network serving as guidance. The implementation of diffusion process in a compact latent space significantly alleviates the massive computational burden typically associated with diffusion models. Importantly, the LHFR synthesized by the DM does not directly intervene in the image restoration process, thereby minimizing the effect of artifacts introduced by DM. Meanwhile the high-frequency details of HQ images and real distribution information encoded in the LHFR substantially boost the stereo image restoration performance, exerting the distribution estimation ability of DM. Extensive experiments on super-resolution, deblurring and low-light enhancement demonstrate that DiffStereo achieves superior results in both accuracy and perceptual quality compared with SOTA methods.

% \begin{acks}
% This work was supported by the National Natural Science Foundation of China (Nos.62171251\&62311530100) and the
% Special Foundation for the Development of Strategic Emerging
% Industries of Shenzhen (JCYJ20200109143010272).
% \end{acks}

%%
%% The acknowledgments section is defined using the "acks" environment
%% (and NOT an unnumbered section). This ensures the proper
%% identification of the section in the article metadata, and the
%% consistent spelling of the heading.
% \begin{acks}
% To Robert, for the bagels and explaining CMYK and color spaces.
% \end{acks}

%%
%% The next two lines define the bibliography style to be used, and
%% the bibliography file.
\bibliographystyle{ACM-Reference-Format}
\bibliography{sample-base}

\end{document}